\documentclass[final]{cvpr}

\usepackage{times}
\usepackage{epsfig}
\usepackage{graphicx}
\usepackage{amsmath}
\usepackage{amssymb}
\usepackage{multirow}
\usepackage{booktabs}
\usepackage{enumitem}
\usepackage{subfigure}
\usepackage{afterpage}
\usepackage{pifont}
\usepackage[dvipsnames]{xcolor}
\usepackage[symbol]{footmisc}

\usepackage[pagebackref=true,breaklinks=true,colorlinks,bookmarks=false]{hyperref}

\makeatletter
\def\and{%
  \end{tabular}%
  \hskip 0.79em \@plus.17fil\relax
  \begin{tabular}[t]{c}}
\makeatother

\def\cvprPaperID{10100} 
\def\confYear{CVPR 2021}

\begin{document}

\title{Lips Don't Lie: A Generalisable and Robust Approach to Face Forgery Detection}

\author{\stepcounter{footnote}Alexandros Haliassos\textsuperscript{1,}\thanks{Corresponding author.}\\
\and
Konstantinos Vougioukas\textsuperscript{1}\\
\and
Stavros Petridis\textsuperscript{1}
\and
Maja Pantic\textsuperscript{1,2}
\and
\textsuperscript{1}Imperial College London
\and
\textsuperscript{2}Facebook London
\and
{\tt\small \{alexandros.haliassos14,k.vougioukas,stavros.petridis04,m.pantic\}@imperial.ac.uk}
}
\maketitle

\renewcommand*{\thefootnote}{\arabic{footnote}}
\setcounter{footnote}{0}
\begin{abstract}
Although current deep learning-based face forgery detectors achieve impressive performance in constrained scenarios, they are vulnerable to samples created by unseen manipulation methods. Some recent works show improvements in generalisation but rely on cues that are easily corrupted by common post-processing operations such as compression. In this paper, we propose \textit{LipForensics}, a detection approach capable of both generalising to novel manipulations and withstanding various distortions. LipForensics targets high-level semantic irregularities in mouth movements, which are common in many generated videos. It consists in first pretraining a spatio-temporal network to perform visual speech recognition (lipreading), thus learning rich internal representations related to natural mouth motion. A temporal network is subsequently finetuned on fixed mouth embeddings of real and forged data in order to detect fake videos based on mouth movements without overfitting to low-level, manipulation-specific artefacts. Extensive experiments show that this simple approach significantly surpasses the state-of-the-art in terms of generalisation to unseen manipulations and robustness to perturbations, as well as shed light on the factors responsible for its performance. Code is available on GitHub.\footnote{\url{https://github.com/ahaliassos/LipForensics}}
\end{abstract}

\section{Introduction}

Recent advances in deep generative models, especially Generative Adversarial Networks (GANs) \cite{goodfellow2014generative}, have enabled the use of off-the-shelf models that can produce ultra-realistic fake videos with little human effort or expertise. Face manipulation methods in particular have raised considerable concerns due to their ability to alter a person's identity \cite{li2020celeb, jiang2020deeperforensics, li2020advancing}, expression \cite{thies2016face2face, thies2019deferred, nirkin2019fsgan}, or lip movements \cite{thies2020neural, vougioukas2019realistic, vougioukas2019end} to match the face in a given target video. The misuse of such technology can spread political propaganda, defame individuals, or damage trust in journalism.

\begin{figure}[tb]
\centering
  \centerline{\includegraphics[width=\linewidth]{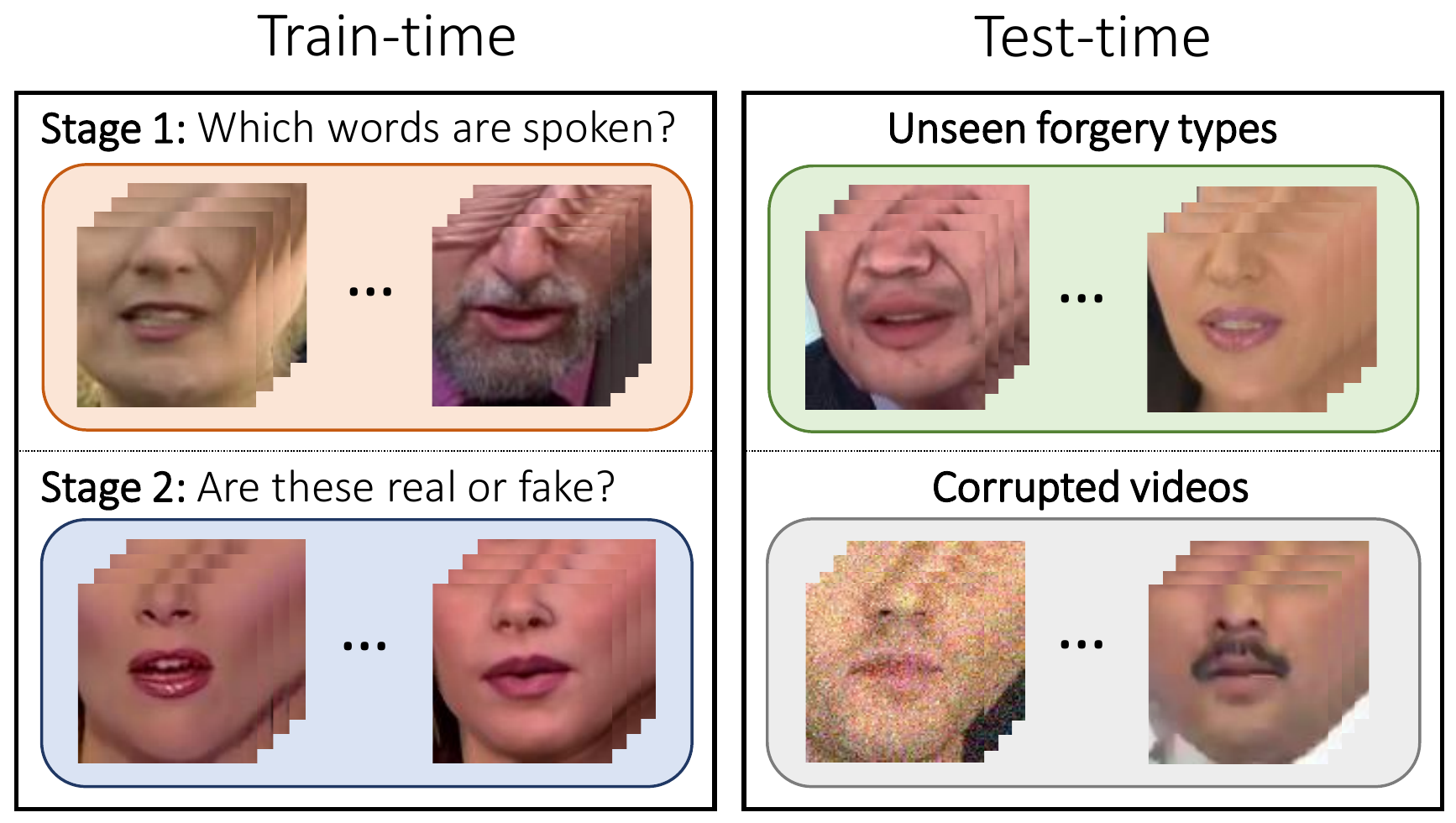}}
 \caption{\textbf{Illustration of our training and testing objectives.} By leveraging semantically high-level, spatio-temporal representations learned via the task of lipreading, our face forgery detector can handle both novel manipulations and common corruptions at test-time.}
 \label{fig:teaser}
\end{figure}

With the aid of recent releases of large-scale face forgery datasets \cite{rossler2019faceforensics++, li2020celeb, jiang2020deeperforensics, dolhansky2020deepfake}, it is possible to train deep convolutional neural networks (CNNs) to detect fake videos \cite{afchar2018mesonet, guera2018deepfake, zhou2017two, agarwal2020detecting, dang2020detection, qian2020thinking, mittal2020emotions, li2018ictu}. Despite excellent accuracy on samples that are independent and identically distributed to the training distribution, it is known that dramatic drops in performance may be experienced when this stringent criterion is not met \cite{li2020face, chai2020makes, nguyen2019multi}. For example, detectors often significantly underperform on novel forgery types. This understandably sparks concerns, as a deployed detector is unlikely to be exposed only to forgeries seen during training. Moreover, they are often sensitive to common perturbations such as compression and are, as a result, vulnerable to the image processing pipelines on social networks.

Recent attempts to boost generalisation to novel forgeries include simple data augmentations \cite{wang2020cnn}, a two-branch network that amplifies multi-band frequencies \cite{masi2020two}, an autoencoder-like structure to serve as an anomaly detector \cite{cozzolino2018forensictransfer, du2019towards, nguyen2019multi}, and patch-based classification to model local patterns \cite{chai2020makes}. However, these methods still substantially overfit to seen manipulations. A particularly effective method is Face X-ray \cite{li2020face}, which proposes to predict the blending boundary between the background image and the inserted, altered face. Although it attains impressive generalisation in cross-manipulation experiments, it relies on often imperceptible patterns which are susceptible to low-level post-processing operations.

It is natural to ask: Are there \textit{semantically high-level} inconsistencies across manipulation methods, which are thus more robust to routine perturbations? We observe that most face video forgeries alter the mouth in some way, to match it with someone else's identity, speech, or expression. Due to the intricate motion of the mouth, current manipulation methods find it difficult to generate movements without falling into the ``uncanny valley\footnote{The ``uncanny valley'' refers to the unease experienced by humans when observing a realistic computer-generated face.}.'' For example, fake mouths often do not adequately close when they pronounce certain phonemes \cite{agarwal2020detecting}. We also notice unnatural fluctuations in the speed of movements, as well as alterations in the shape of the mouth or its interior (\textit{e.g.}, teeth) from frame-to-frame, even when there is no speech (see Appendix \ref{subsec:mouth_ex} for examples). These irregularities provide a precious opportunity for detectors to capitalise on; yet, to the best of our knowledge, previous works do not specifically target mouth motion using spatio-temporal neural networks.

The human visual system can perceive subtle anomalies in forged oral movements as a result of extensive experience in observing real mouths move. In order to endow a network with such experience, we propose to pretrain it on a large corpus of real videos to perform the difficult task of visual speech recognition, also known as lipreading. To be capable of disambiguating similar words, the network must learn rich spatio-temporal representations related to the mouth as well as the teeth and tongue \cite{chung2016lip}. Next, we transfer the acquired knowledge to face forgery detection. Crucially, we treat the first part of the network as a frozen feature extractor that outputs an embedding per frame, and only finetune the temporal convolutional network that takes these embeddings as input. This prevents the network from learning to discriminate the data based on unstable, low-level patterns that may not be generated by other forgery methods.

We dub our approach \textit{LipForensics} (see Figure \ref{fig:teaser}). We conduct extensive experiments to compare its performance with the state-of-the-art in various challenging scenarios. We find that, in most cases, it significantly outperforms previous methods with respect to generalisation to unseen forgeries, while exhibiting remarkable robustness to common corruptions which degrade other models' performance. Further, in-distribution experiments reveal that LipForensics can effectively learn even on heavily compressed data, unlike other detectors. Finally, we validate our design choices through ablation studies, and compare with other large-scale pretraining tasks to demonstrate the superiority of lipreading for achieving generalisable and robust face forgery detection.

\section{Related Work}

\begin{description}[wide,itemindent=\labelsep]
\item[Face forgery detection.] Some earlier face forgery detection works bias the network away from learning high-level features by constraining CNN filters \cite{bayar2016deep} or using relatively shallow networks \cite{afchar2018mesonet} to focus on mesoscopic features. R{\"o}ssler \etal \cite{rossler2019faceforensics++} showed, however, that these are outperformed by a deep, unconstrained Xception \cite{chollet2017xception} network. Some methods \cite{bappy2017exploiting, tarasiou2020extracting} predict the manipulated region along with the label in a multi-task fashion. Another line of research relates to the observation that fully CNN-generated images may exhibit anomalous patterns in the frequency spectrum \cite{frank2020leveraging, durall2020watch, zhang2019detecting, qian2020thinking}. In \cite{guera2018deepfake, sabir2019recurrent}, a CNN followed by an LSTM \cite{hochreiter1997long} are used to capture spatio-temporal features. In \cite{wang2020fakespotter}, fake frames are detected based on anomalies in the neuron behaviour of a face recognition network.

Some works exploit the correspondence between the visual and auditory modalities \cite{agarwal2020detecting, mittal2020emotions, chugh2020not, korshunov2018speaker}. For example, \cite{agarwal2020detecting} targets inconsistencies between the ``M'', ``B'', and ''P'' phonemes and their corresponding visemes, requiring voice-to-text transcriptions and viseme-phoneme alignment. We note that, in contrast, our method is visual-only, \textit{i.e.}, does not take as input the audio during training or inference. A concurrent work \cite{yang2020preventing} also exploits lip movements but is a biometric approach, hence requiring a reference video at inference.

\begin{figure*}[tb]
\centering
  \centerline{\includegraphics[width=\linewidth]{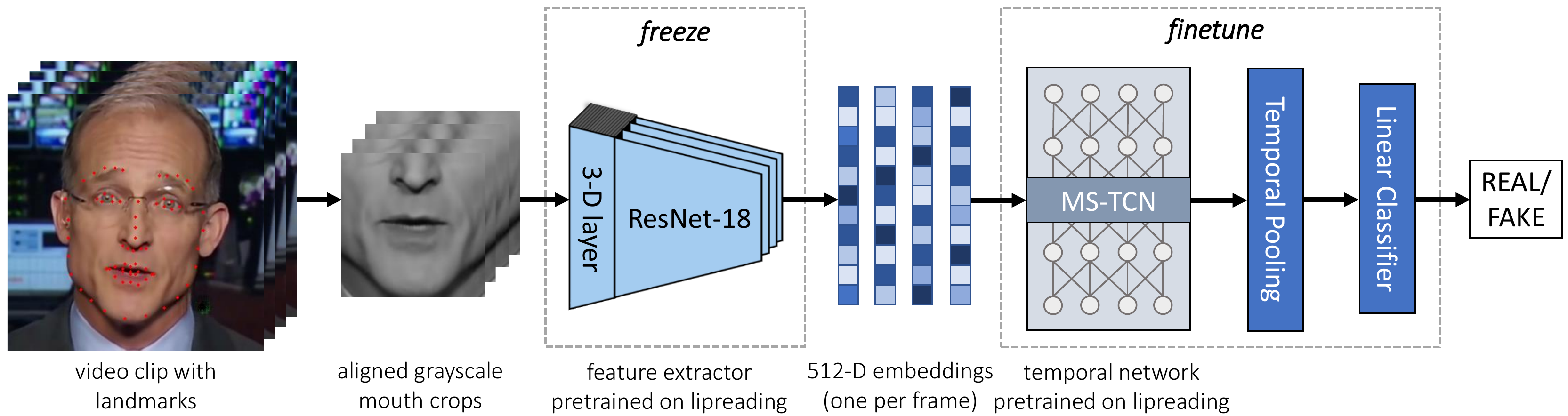}}
 \caption{\textbf{Overview of the finetuning phase on face forgery detection}. The input to the network consists of 25 grayscale, aligned mouth crops (we only show four for illustration purposes). They pass through a frozen feature extractor (a ResNet-18 with an initial 3-D convolutional layer), which has been pretrained on lipreading and hence outputs embeddings sensitive to mouth movements. A multi-scale temporal convolutional network (MS-TCN), also pretrained on lipreading, is finetuned to detect fake videos based on semantically high-level irregularities in mouth motion.}
 \label{fig:overview_finetuning}
\end{figure*}

\item[Generalisation to novel manipulations.]
Although current detectors tend to perform well when the training and test distributions are similar, cross-dataset experiments reveal that they underperform on unseen forgeries \cite{li2020face, chai2020makes, li2020celeb, nguyen2019multi}. To address this issue, ForensicTransfer \cite{cozzolino2018forensictransfer} proposes an autoencoder-like structure to prevent the network from discarding frame information. Follow-up works LAE \cite{du2019towards} and MTDS \cite{nguyen2019multi} additionally exploit binary forgery segmentation masks.  In \cite{masi2020two}, a two-branch recurrent neural network amplifies multi-band frequencies. Wang \etal \cite{wang2020cnn} apply blurring and compression augmentations, which help generalisation across fully-synthetic fake images. Chai \etal \cite{chai2020makes} instead hypothesise that local patterns generalise better and truncate image classifiers to reduce their receptive fields. We challenge this hypothesis by showing that high-level temporal inconsistencies in the mouth region can generalise very well (and lead to more robust detectors).

A particularly successful self-supervised approach is to generate fake videos on-the-fly that contain only the specific irregularities that one may want to target. For instance, FWA \cite{li2019exposing} targets the common affine face warping step in forgery pipelines. More recently, Li \etal \cite{li2020face} observe that many forgery algorithms depend on a blending step to realistically insert an altered face into a background image. They predict the blending boundary between the two images to achieve the current state-of-the-art generalisation on cross-dataset experiments. However, the artefacts that the network focuses on are susceptible to common perturbations, rendering it unsuitable for many real-life scenarios.

\end{description}

\section{LipForensics}
\subsection{Overview}
We tackle the problem of building a generalisable and robust face forgery classifier by distinguishing between natural and anomalous mouth movements. We hypothesise that irregularities in mouth motion exist in fake videos regardless of the generation method that produced them. Due to the semantically high-level nature of these cues, they are also less easily corrupted by common perturbations. However, a spatio-temporal CNN simply trained on mouth crops will not necessarily learn the desired features, as it may separate the data based on other more manipulation-specific cues. To account for this, we take a two-step approach. 

First, we pretrain a CNN, consisting of a spatio-temporal feature extractor followed by a temporal convolutional network, on the task of lipreading. We expect this process to induce internal representations that are sensitive to anomalous dynamics of the mouth in a high-level semantic space, since low-level patterns are likely insufficient for solving
the task. Such pretraining is consistent with recent anomaly detection literature, which suggests that training on the ``normal'' class (in this case real videos) for a suitable task promotes learning of features that are useful for detecting ``anomalous'' samples (in this case fake videos) \cite{golan2018deep, bergman2020goad}. 

Second, we freeze the feature extractor and finetune only the temporal network on forgery data; otherwise, the network may still learn to rely on unwanted artefacts rather than mouth movements. Other works attempt to alleviate overfitting to unstable cues by blurring or adding noise to the input \cite{xuan2019generalization, wang2020cnn}, forcing the network to target blending artefacts \cite{li2020face}, or disincentivising the model from discarding frame information via a reconstruction loss \cite{cozzolino2018forensictransfer, du2019towards, nguyen2019multi}. In contrast, we pass video clips through a deep feature extractor that was trained to perform lipreading and, as a result, outputs embeddings that are relatively invariant to low-level artefacts. We illustrate this process in Figure \ref{fig:overview_finetuning}.

\subsection{Formulation}
Let $\mathcal{X}$ be the set of grayscale video clips, real or fake, centred around the mouth, and let $\mathcal{X}_r\subset\mathcal{X}$ denote the set of real clips. We are given a face forgery dataset $\{(x^j_f, y^j_f)\}_{j=1}^{N_f}$ of size $N_f$, where $x^j_f\in\mathcal{X}$ is a video clip and $y^j_f\in\{0, 1\}$ denotes whether the clip is real or fake. We further assume that we are provided with a labelled lipreading dataset $\{(x^j_l, y^j_l)\}_{j=1}^{N_l}$ of size $N_l$, where $x_l^j\in\mathcal{X}_{r}$ is a real video clip of a word utterance and $y^j_l\in\{0,\dots,L-1\}$ is a label specifying which word was spoken from a vocabulary of length $L$.

We wish to first train a multi-class neural network, $f_l$, to perform lipreading on real videos. It comprises a spatio-temporal feature extractor, a temporal net, and a linear classifier, parameterised by $\theta_g$, $\theta_h$, and $\theta_l$, respectively. The parameters of these subnetworks are randomly initialised and then optimised together to minimise the standard cross entropy loss, $\frac{1}{N_l}\sum_{j=1}^{N_l}\mathcal{L}_{\textsubscript{CE}}(f_l(x^j_l), y^j_l;\theta_g,\theta_h,\theta_l)$.

To learn our forgery detector, $f_f$, we transfer $\theta_g$ and $\theta_h$ and replace the classifier with a binary one, parameterised by $\theta_f$. The temporal net is finetuned (and the classifier is trained from scratch) to minimise the binary cross entropy loss, $\frac{1}{N_f}\sum_{j=1}^{N_f}\mathcal{L}_{\textsubscript{BCE}}(f_f(x^j_f), y^j_f;\theta_h,\theta_f)$. The feature extractor is kept fixed during this phase.

\begin{description}[wide,itemindent=\labelsep]
\item[Architecture.] The feature extractor is a ResNet-18 \cite{he2016deep} with an initial 3-D convolutional layer which preserves the temporal dimension via padding. The feature extractor outputs a 512-D vector for each input frame. The temporal net is a multi-scale temporal convolutional network (MS-TCN) \cite{martinez2020lipreading}, combining short- and long-term temporal information at every layer by concatenating outputs of multiple branches, each with a different temporal receptive field. After a temporal global average pooling layer, a task-specific linear classifier outputs the estimated class probabilities. Architecture details are in Appendix \ref{subsec:arch_details}.

\item[Lipreading pretraining.] The model is pretrained on Lipreading in the Wild (LRW) \cite{chung2016lip}, a dataset containing over 500,000 utterances spanning hundreds of speakers in various poses. It is trained using the approach proposed in \cite{ma2020towards}, which employs born-again distillation \cite{Furlanello2018BornAN}. We use the student classifier in the third generation of teacher-student training. Unless stated otherwise, we use the publicly available, pretrained model found here\footnote{\url{https://github.com/mpc001/Lipreading_using_Temporal_Convolutional_Networks}}.

\item[Preprocessing.] The faces are detected and then aligned to the mean face, after which a scaled $96\times 96$ region around the mouth is cropped and transformed to grayscale. Our clips comprise 25 frames; thus, following random cropping, our input tensor is of size $25\times 88\times 88\times 1$. Note that this size corresponds to a similar number of entries as the standard RGB frame input to many of the forgery detectors in the literature (commonly of size $1\times 256\times 256\times 3$) \cite{li2020face}. More preprocessing details are in Appendix \ref{subsec:preprocessing}. We consider the effect of using the full face as input rather than mouth crops in Appendix \ref{subsec:full_vs_mouth}.

\item[Training the forgery detector.] We use a batch size of 32 and Adam \cite{kingma2014adam} optimisation with a learning rate of $2\times 10^{-4}$. To address any data imbalance, we oversample the minority class. We terminate training when there is negligible improvement to the validation loss for 10 epochs. As data augmentation, we randomly crop the clips with size $88 \times 88$ and horizontally flip with probability 0.5.
\end{description}

\section{Experiments}
\subsection{Setup}
\begin{description}[wide,itemindent=\labelsep]
\item[Datasets.] As in \cite{li2020face, masi2020two}, we mostly use  \textbf{FaceForensics++} (FF++) \cite{rossler2019faceforensics++} as our training dataset due to its forgery diversity. It contains 1.8 million manipulated frames and 4,000 fake videos generated using two face swapping algorithms, DeepFakes \cite{deepfakes} and FaceSwap \cite{faceswap}, and two face reenactment methods, Face2Face \cite{thies2016face2face} and NeuralTextures \cite{thies2019deferred}. As recommended in \cite{rossler2019faceforensics++}, we only use the first 270 frames for each training video, and the first 110 frames for each validation/testing video. Other datasets in our experiments include \textbf{DeeperForensics} \cite{jiang2020deeperforensics} and \textbf{FaceShifter} \cite{li2020advancing}, each featuring an improved face swapping algorithm applied to the real videos from FF++. Further, we use the test set of the face swapping \textbf{Celeb-DF-v2} \cite{li2020celeb} dataset. Finally, we use 3,215 test set videos from the \textbf{DeepFake Detection Challenge} (DFDC) \cite{dolhansky2020deepfake}, where subjects were filmed in extreme conditions, such as large poses and low lighting. More information is in Appendix \ref{subsec:datasets}.

\item[Metrics.] We report results using accuracy (following \cite{rossler2019faceforensics++}) and/or area under the receiver operating characteristic curve (AUC; following \cite{li2020celeb, li2020face}). Since most existing models use a single frame as input, we compute video-level measures for fair comparison: we average the model predictions (each prediction corresponding to a frame or a video clip) across the entire video, as in \cite{masi2020two}. As a result, \textit{all models' video predictions are based on the same number of frames}.
\end{description}

\subsection{Generalisation to unseen manipulations}
Given the rapid advancements in forgery generation, we desire a detector that can correctly classify samples from novel manipulation methods after it is trained on a set of known forgeries. We simulate this scenario below.

\begin{description}[wide,itemindent=\labelsep]
\item[Models for comparison.] For comparison on our experiments using video-level metrics, we train various state-of-the-art detectors designed to improve cross-dataset generalisation as well as some other popular baselines, including: (1) \textbf{Face X-ray} \cite{li2020face}: we train an HRNet-W48 \cite{sun2019deep} both with constructed blended images and fake samples from the considered datasets. (2) \textbf{CNN-aug} \cite{wang2020cnn}: we employ a ResNet-50 with JPEG compression and Gaussian blurring augmentations, both with probability of 0.1. (3) \textbf{Patch-based} \cite{chai2020makes}: we train an Xception classifier truncated after block 3 and average the predictions across the patches. We align the frames to remove rotation variation. (4) \textbf{Xception} \cite{rossler2019faceforensics++}: we train the popular Xception baseline. (5) \textbf{CNN-GRU} \cite{sabir2019recurrent}: we train a DenseNet-161 \cite{huang2017densely} followed by a GRU \cite{cho2014learning}, to compare with a temporal model. All these models take as input a single RGB frame of the full face, except for CNN-GRU, which takes a 5-frame RGB clip. All but Patch-based use ImageNet pretrained weights. See Appendix \ref{subsec:baselines} for more details. We also evaluate the effects of (1) training our spatio-temporal detector from random initialisation (``Scratch''), (2) finetuning the whole network after pretraining on LRW (``Ft whole''), and (3) finetuning only the temporal net (``LipForensics''), all of which are trained on grayscale mouth crops.

\begin{table}[tb]
\begin{center}
\resizebox{\linewidth}{!}{
\begin{tabular}{l c c c c c}\hline
\multirow{2}{*}{Method} & \multicolumn{4}{c}{Train on remaining three} \\  
\cmidrule(lr){2-5}
& DF & FS & F2F & NT & \textbf{Avg} \\ \hline
Xception \cite{rossler2019faceforensics++} & 93.9 & 51.2 & 86.8 & 79.7 & 77.9 \\
CNN-aug \cite{wang2020cnn} & 87.5 & 56.3 & 80.1 & 67.8 & 72.9 \\
Patch-based \cite{chai2020makes} & 94.0 & 60.5 & 87.3 & 84.8 & 81.7 \\
Face X-ray \cite{li2020face} & 99.5 & \textbf{93.2} & 94.5 & 92.5 & 94.9 \\
CNN-GRU \cite{sabir2019recurrent} & 97.6 & 47.6 & 85.8 & 86.6 & 79.4 \\
\hline
Scratch & 93.0 & 56.7 & 98.8 & 98.3 & 86.7 \\
Ft whole & 98.4 & 80.4 & 99.4 & \textbf{99.3} & 94.4   \\
LipForensics (ours) & \textbf{99.7} & 90.1 & \textbf{99.7} & 99.1 & \textbf{97.1} \\ \hline
\end{tabular}
}
\end{center}
\caption{\textbf{Cross-manipulation generalisation.} Video-level AUC (\%) when testing on each forgery type of FaceForensics++ HQ after training on the remaining three. The types are Deepfakes (DF), FaceSwap (FS), Face2Face (F2F), and NeuralTextures (NT).}
\label{table:manip_general}
\end{table}
\item[Cross-manipulation generalisation.] To directly assess generalisation to unseen manipulations, without introducing confounders such as variations in pose or illumination, we experiment here with fake data that were created from the same source videos. Specifically, we test on each of the four methods in FF++ after we train on the remaining three. Rather than using raw videos (as in \cite{li2020face}), we use the high quality (HQ) subset of the dataset (as in \textit{e.g.}, \cite{masi2020two, chai2020makes, nguyen2019multi}), where the videos have been processed with a \textit{visually nearly lossless} compression \cite{rossler2019faceforensics++}. This is more in line with the type of videos found on social media. 

Table \ref{table:manip_general} shows that LipForensics achieves excellent generalisation to novel forgeries, surpassing on average most approaches by a large margin. It also outperforms the previous state-of-the-art method, Face X-ray, by 2.2\% AUC, $94.9\%\rightarrow 97.1\%$. FaceSwap (to which learning-based methods tend not to generalise well, possibly due to its distinct visual artefacts) is the only manipulation for which Face X-ray performs better. Nonetheless, our approach manages to reach 90.1\% AUC without imposing such strong \textit{a priori} knowledge (\textit{i.e.}, that there exists a blending boundary in FaceSwap frames).

We also notice that simply training the spatio-temporal network from scratch on mouth crops leads to surprisingly good results on the expression manipulation methods, \textit{i.e.}, Face2Face and NeuralTextures. This is possibly because there are low-level temporal artefacts around the mouth that exist across the forgery types. More importantly, pretraining on lipreading improves performance across all methods, showing that the network now focuses on more transferable forgery evidence. Finally, freezing the feature extractor provides, on average, a further substantial increase in performance, suggesting that the extractor is prone to overfitting even after pretraining.

\item[Generalisation to other datasets.] To evaluate cross-dataset generalisation, we test a \textit{single model} on DeeperForensics, Faceshifter, Celeb-DF-v2, and DFDC after training on FF++ (all four methods). In Table \ref{table:cross_dataset}, we present the results for the same baselines as well as for three other methods: (1) \textbf{Multi-task} \cite{nguyen2019multi}, which uses an autoencoder-like architecture similar to \cite{cozzolino2018forensictransfer, du2019towards}; (2) \textbf{DSP-FWA} \cite{li2019exposing}; and (3) \textbf{Two-branch} \cite{masi2020two}. We train Multi-task ourselves on FaceForensics++, report  the (video-level) results from the paper of Two-branch \cite{masi2020two}, and evaluate publicly-available, pretrained models for the remaining two methods. 

LipForensics surpasses all methods on every dataset, with especially strong results on FaceShifter and DeeperForensics; this provides promise for its efficacy on future, improved forgeries. All methods obtain relatively low scores on DFDC (and to a lesser extent Celeb-DF), which we attribute to the domain shift caused by significantly different filming conditions. The gains from lipreading pretraining (+8.4\% on average) and from freezing the feature extractor (+4.9\%) are again apparent, providing additional strong evidence that high-level features related to mouth movements are key to generalisation.

\begin{table}
\begin{center}
\resizebox{\linewidth}{!}{
\begin{tabular}{l c c c c c}\hline
Method & CDF & DFDC & FSh & DFo & \textbf{Avg}  \\ \hline
Xception \cite{rossler2019faceforensics++} & 73.7 & 70.9 & 72.0 & 84.5 & 75.3  \\
CNN-aug \cite{wang2020cnn} & 75.6 & 72.1 & 65.7 & 74.4 & 72.0 \\
Patch-based \cite{chai2020makes} & 69.6 & 65.6 & 57.8 & 81.8 & 68.7 \\
Face X-ray \cite{li2020face} & 79.5 & 65.5 & 92.8 & 86.8 & 81.2 \\ 
CNN-GRU \cite{sabir2019recurrent} & 69.8 & 68.9 & 80.8 & 74.1 & 73.4 \\
Multi-task \cite{nguyen2019multi} & 75.7 & 68.1 & 66.0 & 77.7 & 71.9 \\
DSP-FWA \cite{li2019exposing} & 69.5 & 67.3 & 65.5 & 50.2 & 63.1 \\
Two-branch \cite{masi2020two} & 76.7 & --- & --- & --- & --- \\ \hline
Scratch & 62.5 & 65.5 & 84.7 & 84.8 & 74.4 \\
Ft whole & 70.7 & 70.9 & 93.9 & 95.7 & 82.8 \\
LipForensics (ours) & \textbf{82.4} & \textbf{73.5} & \textbf{97.1} & \textbf{97.6} & \textbf{87.7} \\ \hline 
\end{tabular}
}
\end{center}
\caption{\textbf{Cross-dataset generalisation.} Video-level AUC (\%) on Celeb-DF-v2 (CDF), DeepFake Detection Challenge (DFDC), FaceShifter HQ (FSh), and DeeperForensics (DFo) when trained on FaceForensics++.}
\label{table:cross_dataset}
\end{table}

\begin{figure*}[tb]
\centering
  \centerline{\includegraphics[width=\linewidth]{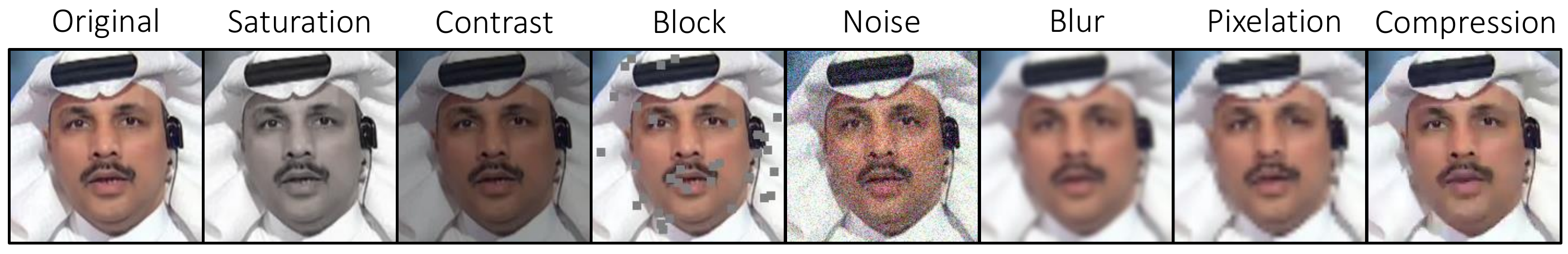}}
 \caption{\textbf{Corruption examples.} Examples of the corruptions considered in our robustness experiments at severity level 3; this set of corruptions was introduced in \cite{jiang2020deeperforensics}. It consists of changes in saturation and contrast, block-wise distortions, white Gaussian noise, Gaussian blurring, pixelation, and video compression. More examples and information can be found in Appendix \ref{subsec:robustness}.}
 \label{fig:corruption_examples}
\end{figure*}

\begin{figure*}[tb]
\centering
  \centerline{\includegraphics[width=\linewidth]{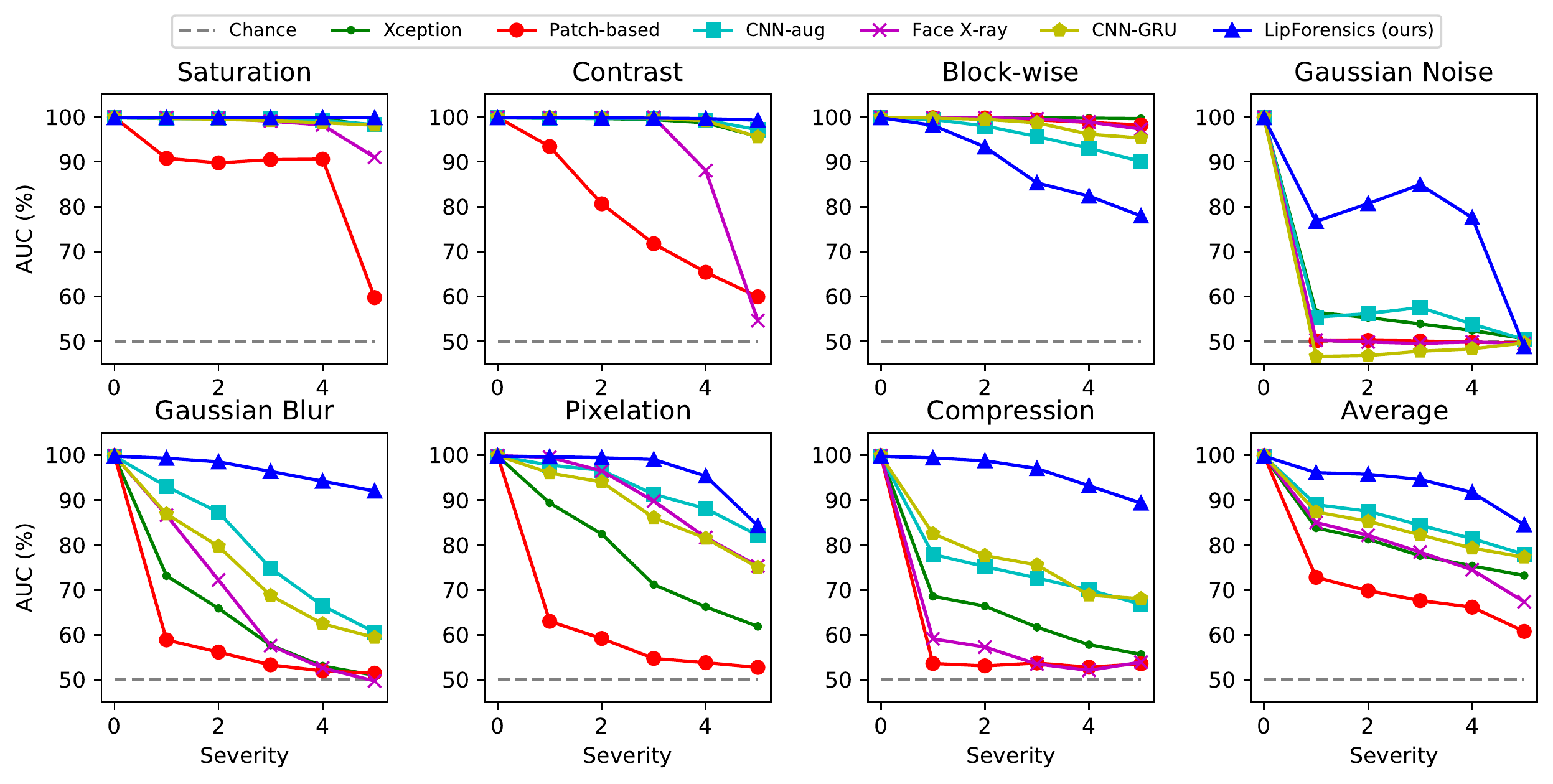}}
 \caption{\textbf{Robustness to various unseen corruptions.} Video-level AUC scores as a function of the severity level for various corruptions. ``Average'' denotes the mean across all corruptions at each severity level. LipForensics is more robust than previous methods to all corruptions except block-wise distortions, which affect high-level semantic content.}
 \label{fig:corruption_severity}
\end{figure*}

\begin{table*}[tb]
\begin{center}
\begin{tabular}{l c c c c c c c c c}\hline
Method & \textcolor{gray}{Clean} & Saturation & Contrast & Block & Noise & Blur & Pixel & Compress & \textbf{Avg} \\ \hline
Xception \cite{rossler2019faceforensics++} & \textcolor{gray}{99.8} & 99.3 & 98.6 & \textbf{99.7} & 53.8 & 60.2 & 74.2 & 62.1 & 78.3 \\
CNN-aug \cite{wang2020cnn} & \textcolor{gray}{99.8} & 99.3 & 99.1 & 95.2 & 54.7 & 76.5 & 91.2 & 72.5 & 84.1 \\
Patch-based \cite{chai2020makes} & \textcolor{gray}{99.9} & 84.3 & 74.2 & 99.2 & 50.0 & 54.4 & 56.7 & 53.4 & 67.5  \\
Face X-ray \cite{li2020face} & \textcolor{gray}{99.8} & 97.6 & 88.5 & 99.1 & 49.8 & 63.8 & 88.6 & 55.2 & 77.5  \\
CNN-GRU \cite{sabir2019recurrent} & \textcolor{gray}{99.9} & 99.0 & 98.8 & 97.9 & 47.9 & 71.5 & 86.5 & 74.5 & 82.3 \\
LipForensics (ours) & \textcolor{gray}{99.9} & \textbf{99.9} & \textbf{99.6} & 87.4 & \textbf{73.8} & \textbf{96.1} & \textbf{95.6} & \textbf{95.6} & \textbf{92.5} \\
 \hline
\end{tabular}
\end{center}
\caption{\textbf{Average robustness to unseen corruptions.} Video-level AUC (\%) on FF++ when videos are exposed to various corruptions, averaged over all severity levels. ``Avg'' denotes the mean across all corruptions (and all severity levels).}
\label{table:robustness}
\end{table*}
\end{description}

\subsection{Robustness to unseen perturbations}

Given the ubiquity of image processing operations on social media, it is critical that deployed forgery detectors are not easily subverted by common perturbations. We investigate the robustness of the detectors by training on uncompressed FF++ and then testing on FF++ samples that were exposed to various unseen corruptions. We consider the following operations at five severity levels, as given in \cite{jiang2020deeperforensics}: changes in saturation, changes in contrast, adding block-wise distortions, adding White Gaussian noise, blurring, pixelating, and applying video compression (H.264 codec). Figure \ref{fig:corruption_examples} gives an example of each corruption at severity level 3. We remove the compression and pixelation training augmentations for Face X-ray and replace the Gaussian blurring augmentation for CNN-aug with median blurring, so that no test-time perturbation is seen during training.

In Figure \ref{fig:corruption_severity}, we show the effect of increasing the severity for each corruption. On Table \ref{table:robustness}, we give the average AUC scores across all severities for each corruption. It is evident that LipForensics is significantly more robust to most perturbations than other methods. For corruptions that affect the high frequency content of the frames (blur, pixelation, compression), it maintains high performance at all but the most severe levels, while the other methods undergo significant deterioration in performance. Patch-based is particularly vulnerable to most corruptions; we attribute this to its reliance on a limited receptive field. Despite its good cross-manipulation performance, Face X-ray is adversely affected by most perturbations, especially compression, suggesting that the blending boundary is easily corruptible. It is also interesting that training with \textit{median} blurring and \textit{JPEG} compression augmentations (CNN-aug) is not an adequate remedy against \textit{Gaussian} blurring and \textit{video} compression. This agrees with CNN robustness literature \cite{geirhos2018generalisation, vasiljevic2016examining}. Finally, LipForensics is sensitive to block-wise distortions, which the other methods are relatively unaffacted by. This result bolsters our hypothesis that other methods tend to focus on low-level cues, whereas LipForensics targets high-level inconsistencies, which block-wise distortions can destroy through occlusion. We also stress that such a corruption is too conspicuous to be used adversarially and is also unlikely to be encountered in practice.

\subsection{Learning on compressed data}
\begin{table}[tb]
\begin{center}
\resizebox{\linewidth}{!}{
\begin{tabular}{l c c c c c c}\hline
\multirow{2}{*}{Method} & \multicolumn{3}{c}{Video-level acc (\%)} & \multicolumn{3}{c}{Video-level AUC (\%)} \\ \cmidrule(lr){2-4} \cmidrule(lr){5-7}
 & Raw & HQ & LQ & Raw & HQ & LQ \\ \hline 
Xception \cite{rossler2019faceforensics++} & 99.0 & 97.0 & 89.0 & 99.8 & 99.3 & 92.0 \\
CNN-aug \cite{wang2020cnn} & 98.7 & 96.9 & 81.9 & 99.8 & 99.1 & 86.9  \\
Patch-based \cite{chai2020makes} & \textbf{99.3} & 92.6 & 79.1 & \textbf{99.9} & 97.2 & 78.3 \\
Two-branch \cite{masi2020two} & --- & --- & --- & --- & 99.1 & 91.1 \\
Face X-ray \cite{li2020face} & 99.1 & 78.4 & 34.2 & 99.8 & 97.8 & 77.3 \\
CNN-GRU \cite{sabir2019recurrent} & 98.6 & 97.0 & 90.1 & \textbf{99.9} & 99.3 & 92.2 \\
LipForensics (ours) & 98.9 & \textbf{98.8} & \textbf{94.2} & \textbf{99.9} & \textbf{99.7} & \textbf{98.1} \\ \hline
\end{tabular}
}
\end{center}
\caption{\textbf{Learning on compressed data.} Performance on FF++ when trained and tested on uncompressed (Raw), slightly compressed (HQ), and heavily compressed (LQ) videos.}
\label{table:ff++}
\end{table}
Next, we investigate in-distribution performance on FF++ at different levels of video compression. While we previously studied robustness to \textit{unseen} corruptions, here we \textit{train and test} a separate model for each of the video qualities provided by the dataset: (1) uncompressed videos (raw), (2) compressed videos at high quality (HQ), and (3) compressed videos at low quality (LQ) \cite{rossler2019faceforensics++}. We also report, where applicable, results from the Two-branch paper \cite{masi2020two}.

Table \ref{table:ff++} shows that while all models perform almost flawlessly on raw data, their efficacy varies when trained on compressed videos. Methods that employ deep networks (\textit{e.g.}, Xception and CNN-aug) reach relatively high performance on the LQ dataset. In contrast, patch-based classification struggles to effectively discriminate the data. We also notice that Face X-ray, albeit employing a very deep network, suffers the most under compression. This empirically validates our intuition that the blending artefacts are largely destroyed when the videos are compressed. On the other hand, LipForensics is substantially less affected by compressed data, outperforming all other methods, as it targets high-level spatio-temporal cues. We believe that the improvements over CNN-GRU and Two-branch, which also account for the temporal dimension, are due to the lipreading pretraining, the different architectures (convolutional versus recurrent), and the more frames per clip.

\subsection{Ablation study} \label{sec:ablations}

\begin{figure}[tb]
\centering
  \centerline{\includegraphics[width=\linewidth]{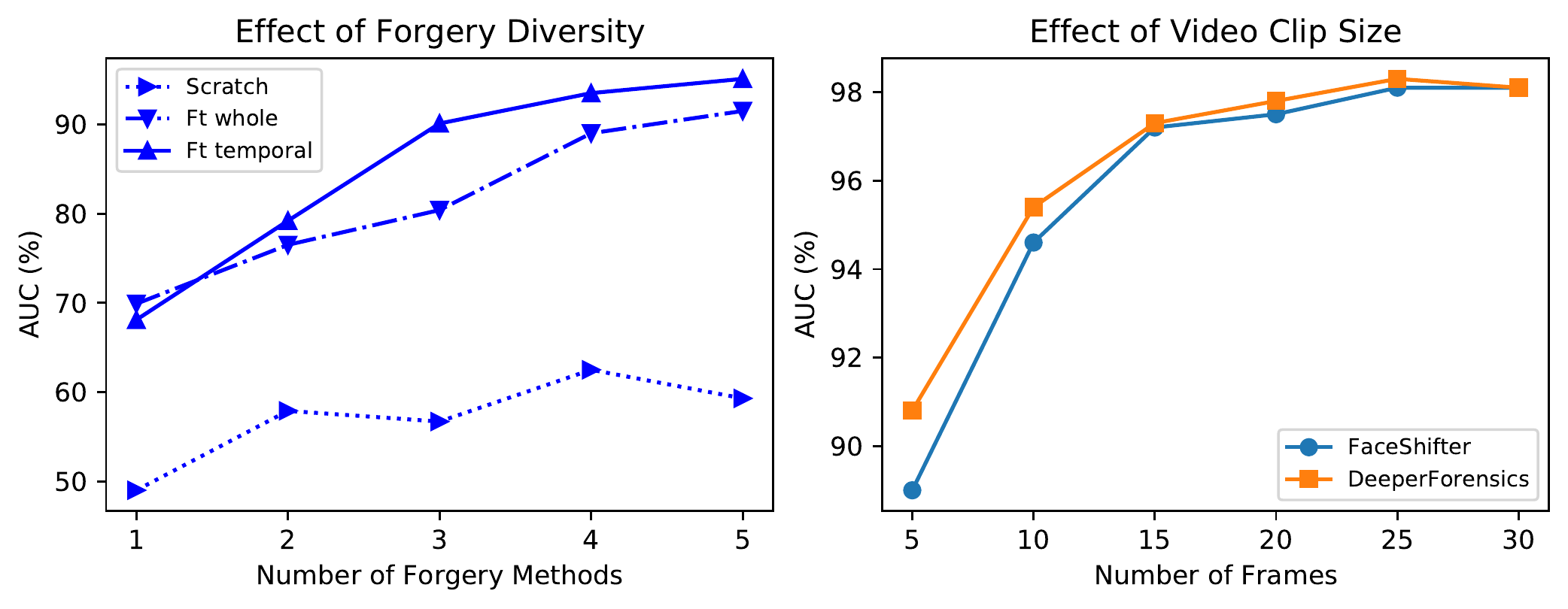}}
 \caption{\textbf{Model analysis.} Effect of increasing forgery diversity on generalisation to FaceSwap (\textbf{left}; details in Section \ref{sec:ablations}). Effect of video clip size on generalisation to FaceShifter and DeeperForensics when trained on FF++ (\textbf{right}).}
 \label{fig:model_analysis}
\end{figure}

\begin{description}[wide,itemindent=\labelsep]
\item[Effect of increasing forgery diversity.]
With the development of new forgery generators, it will be possible to increase the diversity of manipulation methods in the training set. We investigate how this affects generalisation. We treat FaceSwap as the unseen method, as it appears to be the most difficult for our detector to generalise to (see Table \ref{table:manip_general}). We first train only on Deepfakes and then cumulatively add the following: NeuralTextures, FaceSwap, FaceShifter, and DeeperForensics (which share the same source videos).

In Figure \ref{fig:model_analysis}, we see that LipForensics scales well with the forgery diversity, reaching 95.1\% AUC on FaceSwap. Notably, finetuning the whole network still has inferior performance to freezing the feature extractor, even when training on five forgery methods. Without lipreading pretraining, the model does not achieve adequate generalisation.
\item[Influence of video clip size.] We next study the effect of video clip size on generalisation, when trained on FF++ (HQ). For the number of frames in each video clip, we consider the set $\{5, 10, 15, 20, 25, 30\}$ and train a new model for each clip size. As Figure \ref{fig:model_analysis} shows, increasing the number of frames tends to improve performance. A size of 25 strikes a good balance between performance and computational/memory cost. 

\begin{table}
\begin{center}
\resizebox{\linewidth}{!}{
\begin{tabular}{l c c c c}\hline
Model & Finetune & FSh & DFo  \\ \hline
ResNet-18 & whole & 59.3 & 75.7 \\
ResNet-3D/2D & whole & 67.1 & 74.6 \\
ResNet-3D/2D+MS-TCN & whole & 83.2 & 84.6 \\ 
ResNet-3D/2D+MS-TCN (ours) & temporal & \textbf{87.5} & \textbf{90.4} \\ \hline
\end{tabular}
}
\end{center}
\caption{\textbf{Effect of different components.} All models are pretrained on LRW. We report video-level accuracy (\%) scores on FaceShifter (FSh) and DeeperForensics (DFo) when finetuned on FaceForensics++. Last row corresponds to LipForensics.}
\label{table:dif_components}
\end{table}

\item[Effect of different components.] We further study the importance of modelling high-level temporal inconsistencies. We pretrain on LRW a ResNet-18 (without a 3-D layer) followed by an MS-TCN, and then finetune only the ResNet-18 on FF++; thus, no temporal information can be exploited. We compare it with finetuning the ResNet-3D/2D, which can capture short-term, low-level temporal dynamics, and with the full spatio-temporal model that can model long-term, high-level irregularities. On Table \ref{table:dif_components}, we see that modelling short-term dynamics improves generalisation to FaceShifter but not DeeperForensics. More importantly, \textit{ high-level} temporal information is crucial for generalisation, as indicated by the large improvements when the MS-TCN is added and also when the ResNet is kept frozen.

\begin{table}
\begin{center}
\resizebox{\linewidth}{!}{
\begin{tabular}{l c c c c}\hline
Model & \# params & Pretrain & FSh & DFo  \\ \hline
R(2+1)D-18 & 31.3M & none & 63.6 & 65.4 \\
R(2+1)D-18 & 31.3M & Kinetics & 65.7 & 68.2 \\ \hline
ip-CSN-152 & 32.2M & none & 68.2 & 65.7  \\
ip-CSN-152 & 32.2M & Kinetics & 73.9 & 76.4 \\
ip-CSN-152 & 32.2M & IG-65M & 66.1 & 69.6 \\ \hline
SE-ResNet50 & 43.8M & none & 60.0 & 70.7  \\ 
SE-ResNet50 & 43.8M & FR & 64.3 & 68.9 \\ \hline
ResNet+MS-TCN & 36.0M & none & 62.5 & 61.4  \\
ResNet+MS-TCN & 36.0M & LRW & 83.2 & 84.6 \\ 
ResNet+MS-TCN\textsuperscript{*} (ours) & 24.8M & LRW & \textbf{87.5} & \textbf{90.4} \\ \hline
\end{tabular}
}
\end{center}
\caption{\textbf{Other pretraining datasets.} Effect of pretraining on Kinetics, IG-65M, and face recognition (FR) datasets. We give video-level accuracy scores (\%) on FaceShifter and DeeperForensics when finetuned on FF++. ``\# params'' refers to trainable parameters, and asterisk (*) denotes freezing the feature extractor (corresponding to LipForensics). }
\label{table:pretraining}
\end{table}

\item[Other pretraining datasets.] Does pretraining on other large-scale datasets work just as well as pretraining on lipreading? We note that most baselines considered thus far use ImageNet weights. Here, we further consider spatio-temporal models pretrained on Kinetics-400 \cite{kay2017kinetics} and IG-65M \cite{ghadiyaram2019large}. The former dataset comprises around $300,000$ video clips spanning 400 action classes, and the latter is a massive-scale weakly-supervised dataset containing over 65 million social media clips with corresponding hashtags. We use the state-of-the-art R(2+1)D-18 \cite{tran2018closer} and ip-CSN-152 \cite{tran2019video} models, which have a similar number of parameters as the ResNet-18+MS-TCN model we use. Furthermore, we compare with a Squeeze-and-Excitation \cite{hu2018squeeze} ResNet-50, pretrained using ArcFace loss \cite{deng2019arcface} on an amalgamation of face recognition datasets\footnote{\url{https://github.com/TreB1eN/InsightFace_Pytorch}}. 

We finetune all models on FF++ and test on FaceShifter and DeeperForensics to evaluate cross-manipulation generalisation (see Table \ref{table:pretraining}). Although other pretraining tasks generally improve performance over training from scratch, none of them match the generalisation achieved by pretraining on lipreading. This confirms its importance in inducing representations suitable for forgery detection. 

\begin{figure}[tb]
\centering
  \centerline{\includegraphics[width=\linewidth]{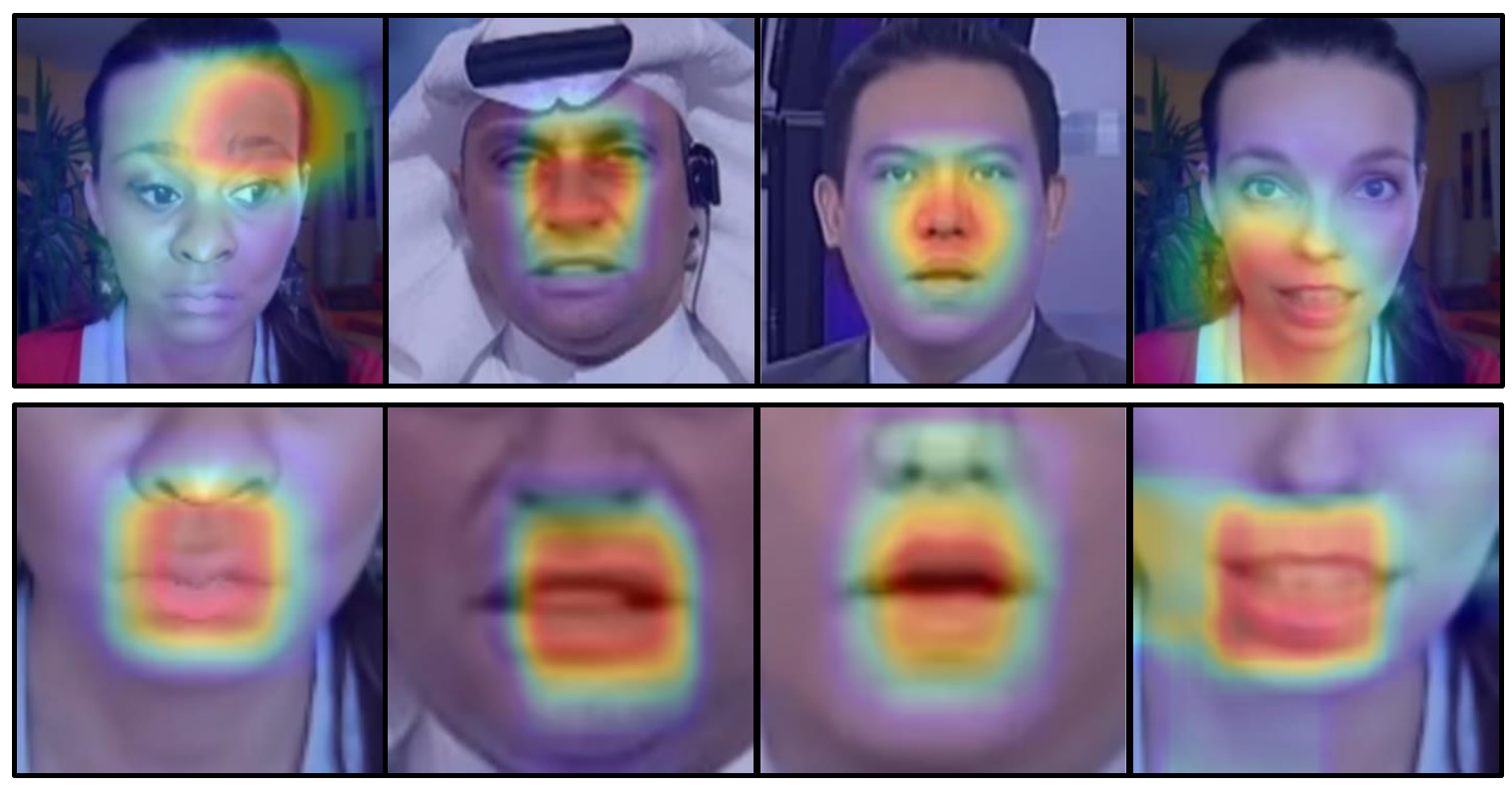}}
 \caption{\textbf{Occlusion sensitivity.} Visualisation of the regions which Xception (\textbf{top}) and LipForensics (\textbf{bottom}) rely on.}
 \label{fig:occlusion_sensitivity}
\end{figure}

\item[Occlusion sensitivity.] To visualise the spatial regions which the detectors rely on, we occlude different portions of the input by sliding a grey square or a square cuboid, and observe the change in the class probability, as proposed in \cite{zeiler2014visualizing}. In Figure \ref{fig:occlusion_sensitivity}, we give results on randomly selected samples for LipForenics and Xception, both trained on FaceForensics HQ. As expected, LipForensics focuses predominantly on the mouth, which is not the case for Xception. More visualisations are in Appendix \ref{subsec:occlusion}.

\end{description}

\subsection{Limitations}
Despite its performance on various experiments, we acknowledge that LipForensics is not without limitations. It cannot be applied to isolated images, and it would probably not detect a fake video in which the mouth is occluded or left unaltered. However, arguably the most pernicious forgery content is videos where the mouth has been manipulated to modify speech, identity, or expression. It is also possible that there is a performance decline with limited mouth motion, although within most 25-frame clips, we noticed at least some movement, even in the absence of speech. Finally, it requires a large-scale labelled dataset for pretraining. Examples of failure cases are given in Appendix \ref{subsec:failures}.

\section{Conclusion}
In this paper, we proposed a novel approach, dubbed \textit{LipForensics}, for the detection of forged face videos. It targets inconsistencies in semantically high-level mouth movements by leveraging rich representations learned via lipreading. It achieves state-of-the-art generalisation to unseen forgery types while being significantly more robust than other methods to various common corruptions. Meeting both of these objectives is crucial for face forgery detection in real-life, and we believe that our work is an important step in the fight against fake videos.

\begin{description}[wide,itemindent=\labelsep]
\item[Acknowledgements.] We would like to thank Pingchuan Ma for his help with lipreading experiments. Alexandros Haliassos was financially supported by an Imperial President's PhD Scholarship. All training, testing, and ablation studies have been conducted at Imperial College London. 
\end{description}

{\small
\bibliographystyle{ieee_fullname}
\bibliography{egbib}
}

\clearpage
\appendix

\begin{figure}[tb]
\centering
  \centerline{\includegraphics[width=\linewidth]{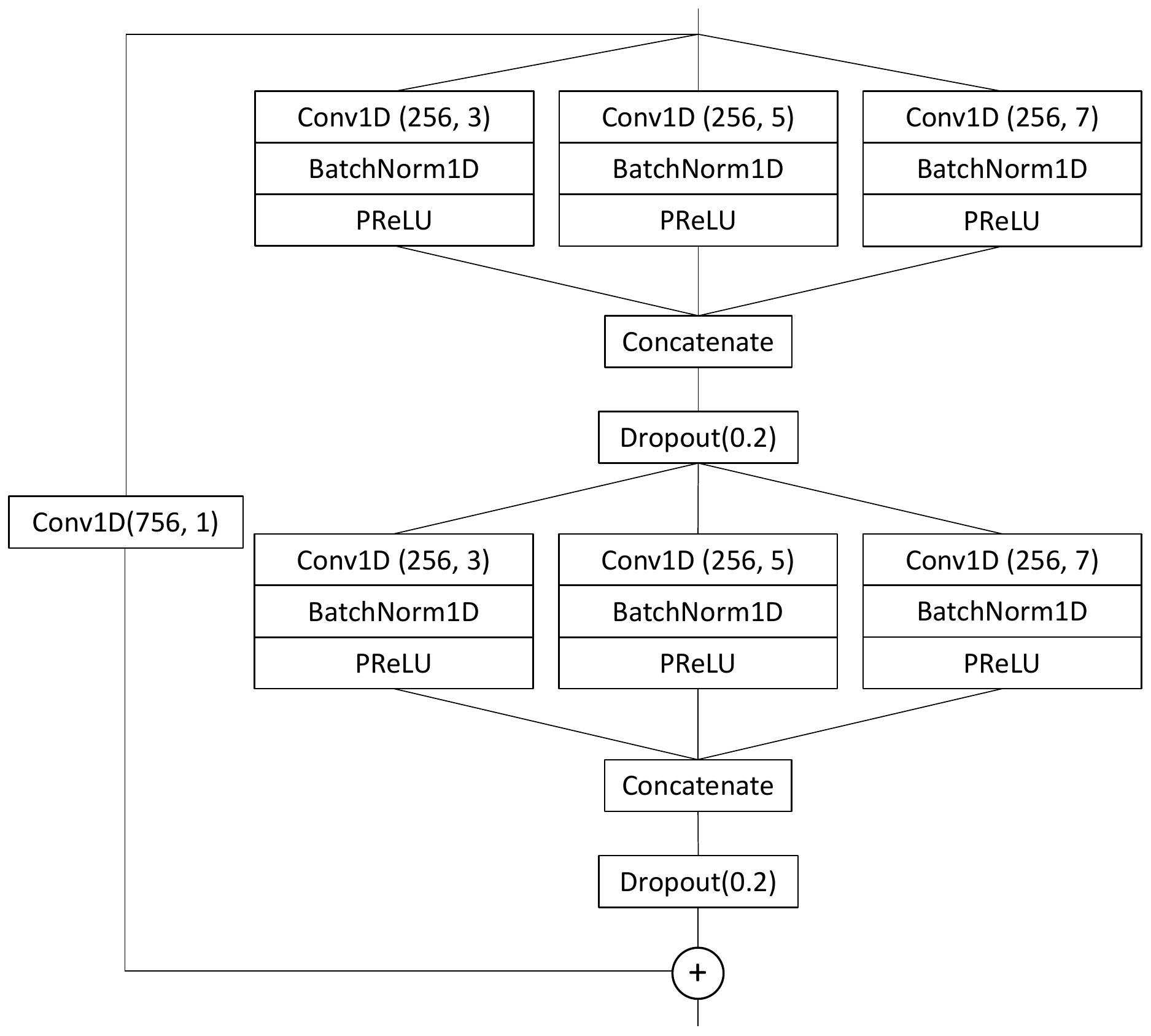}}
\caption{\textbf{Diagram of an MS-TCN block.} Diagram of a single block in the multi-scale temporal convolutional network (MS-TCN) that we employ. The stride is 1 for all convolutions. The full temporal network we use consists of four such blocks. The dilation rate in the first block is equal to 1, and each subsequent block's dilation rate is $2\times$ the previous one's.}
 \label{fig:tcn_block}
\end{figure}

\begin{figure*}[tb]
\centering
  \centerline{\includegraphics[width=\linewidth]{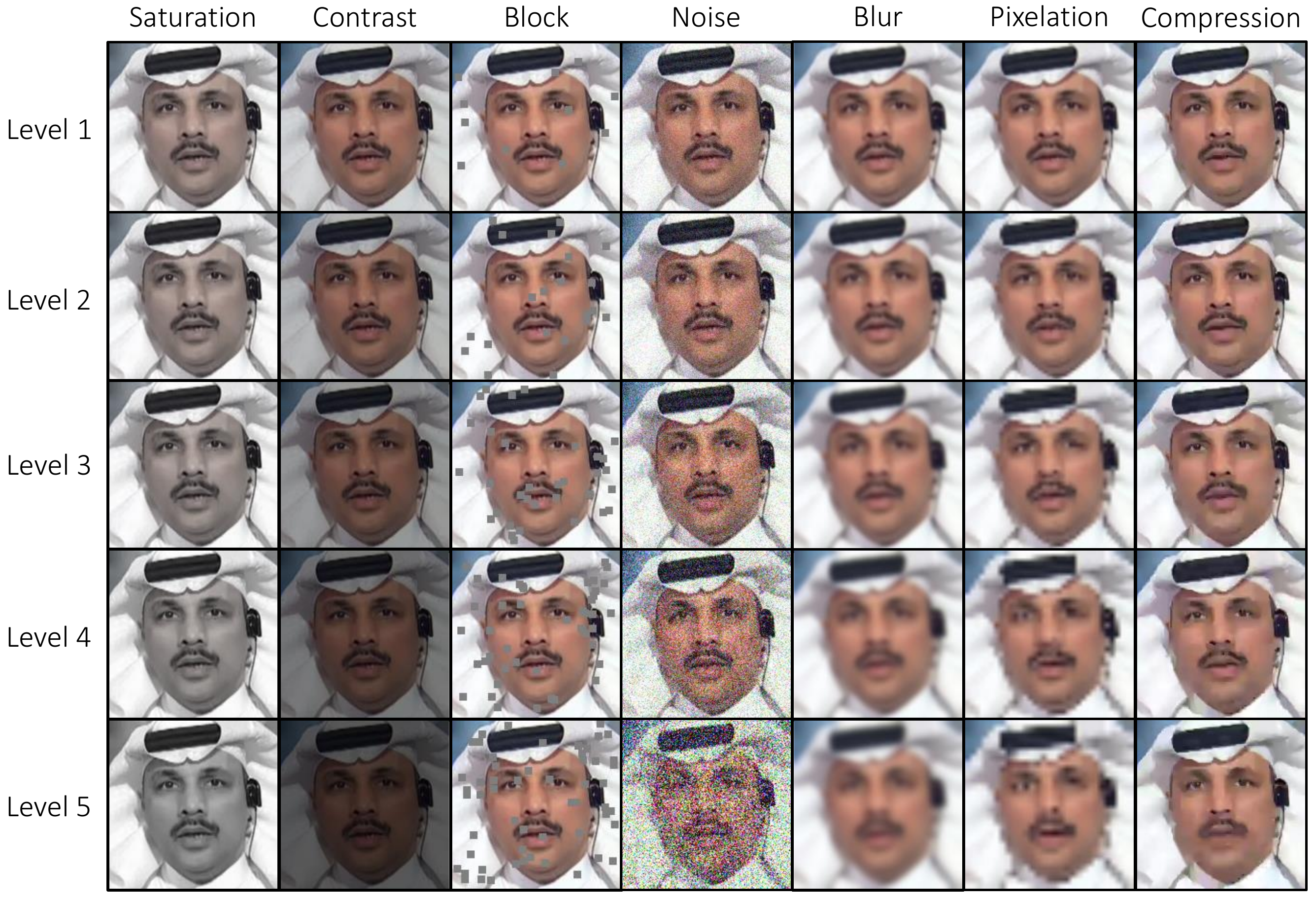}}
\caption{\textbf{All types of perturbations at all severity levels.} Visualisation of the seven perturbation types we used in our robustness experiments at all five of the severity levels. We note that in \cite{jiang2020deeperforensics}, ``Pixelation'' is named ``JPEG compression,'' though the official code, at the time of writing, indeed performs pixelation (downscaling and upscaling).}
 \label{fig:perturbations}
\end{figure*}

\section{More Implementation Details}
\subsection{Architecture details} \label{subsec:arch_details}

A single block of the multi-scale temporal convolutional network \cite{martinez2020lipreading} (MS-TCN) used in our architecture is shown in Figure \ref{fig:tcn_block}. The abbreviations are defined as follows:
\begin{itemize}
    \item $\text{Conv1D}(x, y)$: 1-D convolutional layer with $x$ output channels and kernel size $y$. All use ``same'' padding and stride of 1.
    \item BatchNorm1D: 1-D batch normalisation \cite{ioffe2015batch} with momentum of 0.1.
    \item PReLU: Parametric ReLU activation \cite{he2015delving} with a separate learnable parameter for each input channel.
    \item Dropout($x$): Dropout layer \cite{srivastava2014dropout} with probability $x$.
\end{itemize}

\subsection{Datasets} \label{subsec:datasets}

\begin{description}[wide,itemindent=\labelsep]
\item[FaceForensics++ (FF++) \cite{rossler2019faceforensics++}.] We download the dataset from the official webpage\footnote{\url{https://github.com/ondyari/FaceForensics}}. We use the provided training/validation/test splits.

\item[FaceShifter \cite{li2020advancing}.] We download the FaceShifter samples (at c23 compression) from the same place as FF++, since these have been recently added to the webpage. Note that when we refer to FF++, we are referring to the version described in the FF++ paper, \textit{i.e.}, containing the 4 manipulation methods without FaceShifter. We use the same training/validation/test splits as in FF++.

\item[DeeperForensics \cite{jiang2020deeperforensics}.] We download the dataset from the official webpage\footnote{\url{https://github.com/EndlessSora/DeeperForensics-1.0}}. We use the same training/validation/test splits as in FF++.

\item[Celeb-DF-v2 \cite{li2020celeb}.] We download the dataset from the official webpage\footnote{\url{https://github.com/yuezunli/celeb-deepfakeforensics}}. We use the test set, which consists of 518 videos.

\item[DFDC \cite{dolhansky2020deepfake}.] We download the test set of the full DFDC dataset from the official webpage\footnote{\url{https://ai.facebook.com/datasets/dfdc}}. Some videos feature more than one person. To remove ambiguities in preprocessing, we only use single-person videos. Further, many videos have been filmed in extreme conditions (lighting, poses, etc) and/or have been post-processed with aggressive corruptions. As such, we only use videos for which the face and landmark detectors did not fail.
\end{description}
\subsection{Preprocessing} \label{subsec:preprocessing}
We use RetinaFace \cite{deng2020retinaface}\footnote{\url{https://github.com/biubug6/Pytorch\_Retinaface}} to detect a face for each frame in the videos. As in \cite{rossler2019faceforensics++}, we only extract the largest face and use an enlarged crop, 1.3$\times$ the tight crop produced by the face detector. To crop the mouths for LipForensics, we compute 68 facial landmarks using FAN \cite{bulat2017far}\footnote{\url{https://github.com/1adrianb/face-alignment}}. The landmarks are smoothed over 12 frames to account for motion jitter, and each frame is affine warped to the mean face via five landmarks (around the eyes and nose). The mouth is cropped in each frame by resizing the image and then extracting a fixed $96\times 96$ region centred around the mean mouth landmark. We note that alignment is performed to remove translation, scale, and rotation variations; it does not affect the way the mouth moves.
\subsection{Baselines} \label{subsec:baselines}
For the baselines we consider, we provide details on our implementations that are not given in the main text. Unless stated otherwise, Adam \cite{kingma2014adam} optimisation is used with a learning rate of $2\times 10^{-4}$ and batch size of 32.
\begin{description}[wide,itemindent=\labelsep]
\item[\textbf{Face X-ray} \cite{li2020face}.] To generate the blended images for training, we use provided code\footnote{\url{https://github.com/AlgoHunt/Face-Xray}}. In addition to the random mask deformation and colour correction operations described in the paper, the following augmentations are applied as per the code: random horizontal flipping, JPEG compression (with quality $\sim \text{Uniform}\{30, 31, \dots, 100\}$), and pixelation (downscaling image by a factor $\sim\text{Uniform}[0.2, 1]$), each with probability 0.5. For fair comparison with the other methods, we also train with samples from FaceForensics++ (FF++). Following the code, each image sampled during training is either a real FF++ frame or a fake sample, with probability 0.5. In turn, each fake sample is either a blended image or an FF++ fake frame, again with probability 0.5. The cropped faces are resized to $317\times 317$ and then centre cropped to $256\times 256$. The scaling factor, $\lambda$, corresponding to the segmentation loss is set to 100, as in the paper.

\item[\textbf{CNN-aug} \cite{wang2020cnn}.] We use the official code\footnote{\url{https://github.com/peterwang512/CNNDetection}}. The cropped faces are resized to $256\times 256$. We use JPEG compression (with quality $\sim \text{Uniform}\{60, 61, \dots, 100\}$) and Gaussian blurring with standard deviation $\sim\text{Uniform}[0,3]$, both with probability 0.1. We also use horizontal flipping with probability 0.5.

\item[\textbf{Patch-based} \cite{chai2020makes}.] We use the official code\footnote{\url{https://github.com/chail/patch-forensics}}. We train the model ourselves, since no provided pretrained model was trained on full FF++. The faces are aligned by affine warping them to the mean face and then resized to $299\times 299$. We use horizontal flipping with probability 0.5. Adam \cite{kingma2014adam} with learning rate $1\times 10^{-3}$ is used, as suggested in the paper.

\item[\textbf{Xception} \cite{rossler2019faceforensics++}.] We use the official code\footnote{\url{https://github.com/ondyari/FaceForensics}}. The cropped faces are resized to $299\times 299$. We use horizontal flipping with probability 0.5.

\item[\textbf{CNN-GRU} \cite{sabir2019recurrent}.] The cropped faces are resized to $224\times 224$. We use horizontal flipping with probability 0.5. As recommended in \cite{sabir2019recurrent}, we first train only the DenseNet-161 \cite{huang2017densely} (by adding a linear classifier). We then append a single-layer, bi-directional GRU \cite{cho2014learning} with hidden size 128 and train the whole network end-to-end.

\item[\textbf{Multi-task} \cite{nguyen2019multi}.] We use the official code\footnote{\url{https://github.com/nii-yamagishilab/ClassNSeg}} and follow the paper recommendations for all hyperparameters. We use the ``deep'' version of the model. We train it ourselves since the provided pretrained model has only been trained on a subset of FF++. The cropped faces are resized to $256\times 256$. We use horizontal flipping with probability 0.5. Adam \cite{kingma2014adam} with learning rate $1\times 10^{-3}$ is used, as suggested in the paper.

\item[\textbf{DSP-FWA} \cite{li2019exposing}.] We use the official code\footnote{\url{https://github.com/yuezunli/DSP-FWA}} and pretrained model (on self-collected real faces), which uses a dual spatial pyramid approach. Each face is aligned and extracted at 10 different scales. They are all resized to $224\times 224$. 

\item[\textbf{R(2+1)D-18} \cite{tran2018closer} and \textbf{ip-CSN-152} \cite{tran2019video}.] We use the official code\footnote{\url{https://github.com/facebookresearch/VMZ}} and finetune pretrained models. We perform the same preprocessing as for our LipForensics approach, except that RGB frames are used rather than grayscale, since the pretrained tasks use colour frames. 

\item[\textbf{SE-ResNet50} \cite{hu2018squeeze}.] We use the ArcFace \cite{deng2019arcface} code\footnote{\url{https://github.com/TreB1eN/InsightFace_Pytorch}} and finetune the backbone of the model pretrained on face recognition datasets. The cropped faces are resized to $112\times 112$, since this is the size used during pretraining. We use horizontal flipping with probability 0.5.

\end{description}

\subsection{Robustness experiments} \label{subsec:robustness}
To apply the corruptions in our robustness experiments, we use the DeeperForensics code\footnote{\url{https://github.com/EndlessSora/DeeperForensics-1.0/tree/master/perturbation}}. All considered corruptions at all severity levels are depicted in Figure \ref{fig:perturbations}. 

\section{Full Face Versus Mouth Crops} \label{subsec:full_vs_mouth}

\begin{table}
\begin{center}
\begin{tabular}{l c c c c}\hline
Input type & Pretrain & Finetune & FSh & DFo  \\ \hline
Full face & none & whole & 68.2 & 67.1 \\
Full face & LRW & whole & 82.9 & 85.2 \\
Full face & LRW & temporal & 84.3 & 90.0 \\ \hline
Mouth & none & whole & 62.5 & 61.4 \\
Mouth & LRW & whole & 83.2 & 84.6 \\
Mouth & LRW & temporal & \textbf{87.5} & \textbf{90.4} \\ \hline
\end{tabular}
\end{center}
\caption{\textbf{Full face crops versus mouth crops.} Effect of training on tight full face crops compared with training on mouth crops. We report video-level accuracy (\%) scores on FaceShifter (FSh) and DeeperForensics (DFo) when trained on FaceForensics++.}
\label{table:fullface}
\end{table}

In the main text, we always use mouth crops for LipForensics. Here, we increase the crop from $88\times 88$ to $112\times 112$ (after random cropping) to also include the whole nose and eyes in the input. We pretrain a new model on LRW using this input. As shown in Table \ref{table:fullface}, when training from scratch, using full faces rather than
mouth crops yields better generalisation to FaceShifter and DeeperForensics, but when using lipreading pretraining, mouth crops perform better. For both types of input, lipreading pretraining improves accuracy significantly.

\section{Qualitative Analysis}
\subsection{High-level mouth inconsistencies} \label{subsec:mouth_ex}
\begin{figure*}[t]
\centering
  \centerline{\includegraphics[width=\linewidth]{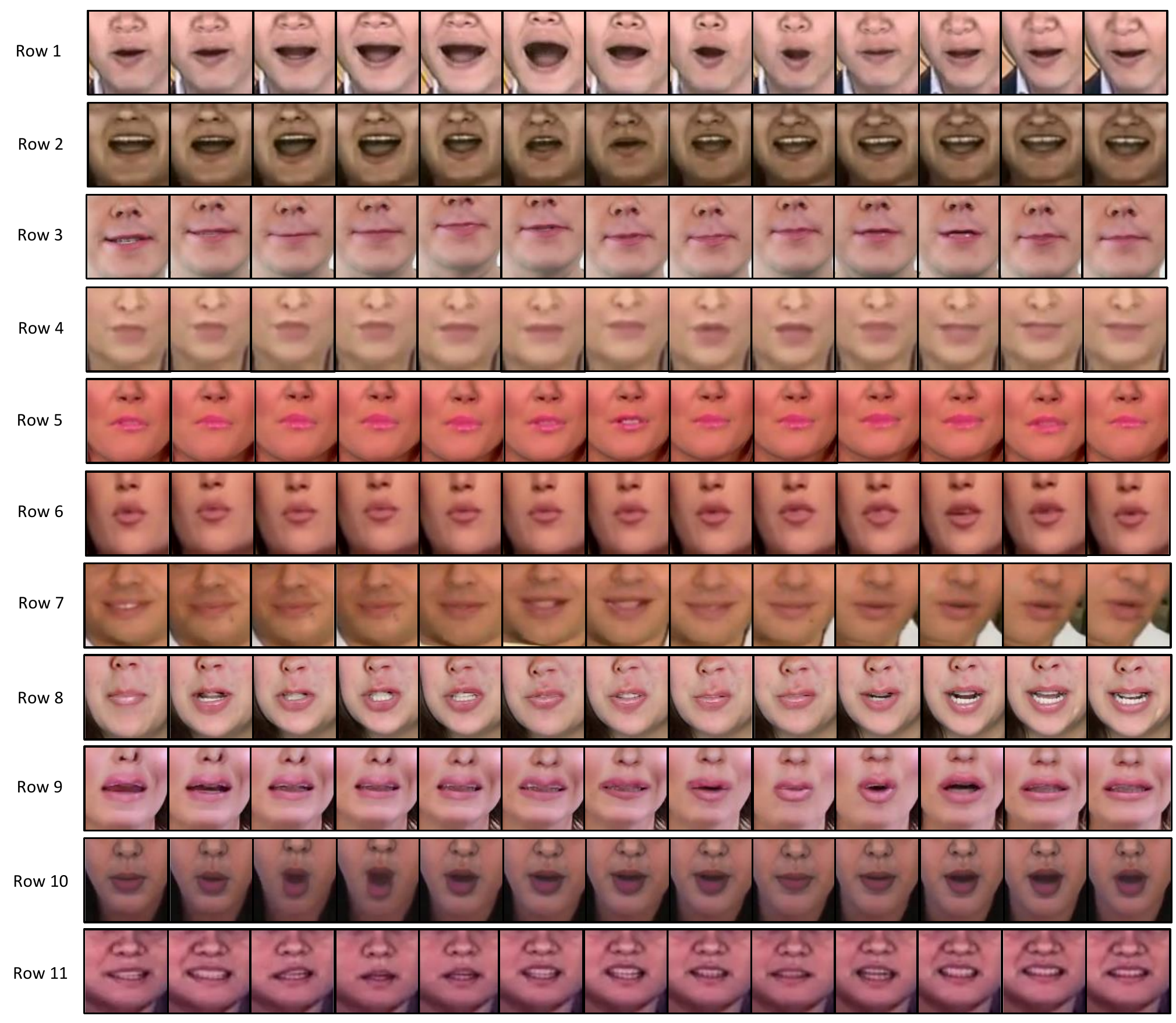}}
\caption{\textbf{Examples of semantically high-level inconsistencies around the mouth region.} Rows 1-2 show mouths that do not sufficiently close; rows 3-7 show mouths with limited mouth movements but which still exhibit anomalous behaviour; rows 8-9 show inconsistencies in the teeth and lip shape; rows 10-11 show temporal irregularities in mouth shape (\textit{e.g.}, see frames 3 and 4 in row 10 and frame 3 in row 11). Subtle anomalies are more readily observed in video form.}
 \label{fig:high_level_inconsistencies}
\end{figure*}

Our approach targets high-level temporal inconsistencies related to the mouth region. We show examples of such anomalies in Figure \ref{fig:high_level_inconsistencies}. Notice that in some cases, the mouth does not sufficiently close, as noted in \cite{agarwal2020detecting}. In other cases, subtle temporal inconsistencies in the shape of the mouth or its interior (\textit{e.g.}, teeth) are present. 

\subsection{Failure cases} \label{subsec:failures}
\begin{figure*}
\centering
  \centerline{\includegraphics[width=\linewidth]{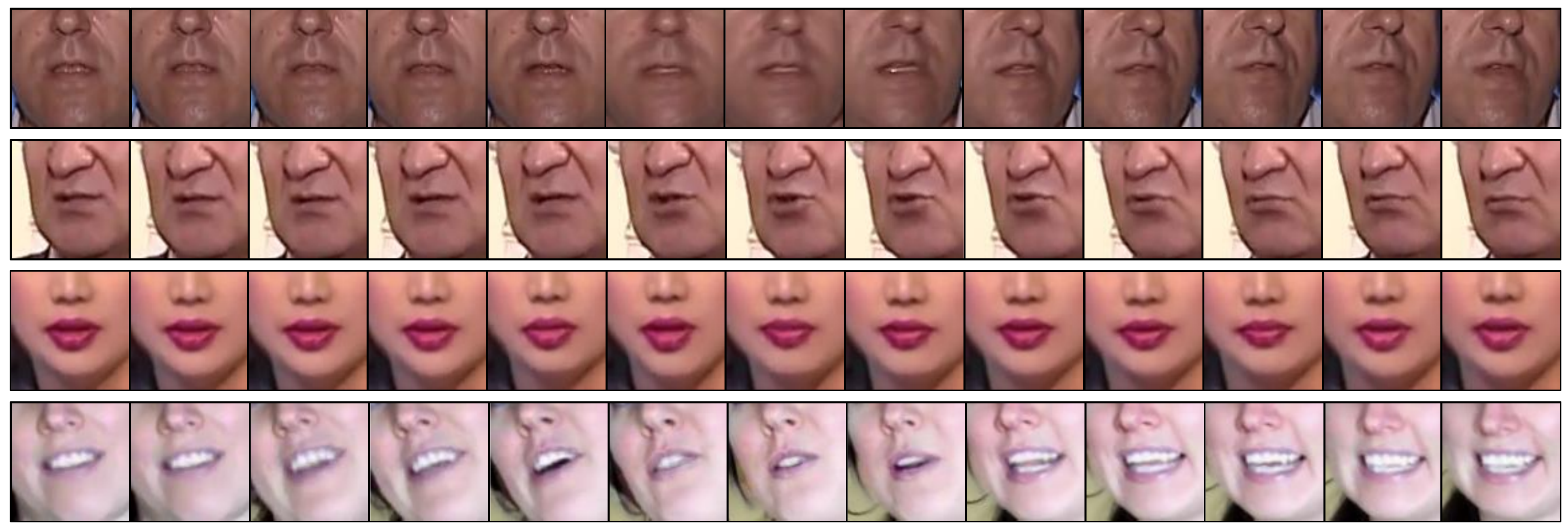}}
\caption{\textbf{Failure cases.} Top two rows are real videos predicted as fake and bottom two are fake videos predicted as real.}
 \label{fig:failure_cases}
\end{figure*}

Examples of failure cases are given in Figure \ref{fig:failure_cases}. In general, we noticed that many of the failure cases involve rapid head movements, poses that are uncommon in the training set (FF++), or very limited mouth movements.

\subsection{Occlusion sensitivity} \label{subsec:occlusion}
We show more visualisation examples using the occlusion sensitivity approach discussed in the main text. This approach was introduced in \cite{zeiler2014visualizing}. It relies on systematically covering up different portions of the frames with a grey block and measuring the effect on the predictions of the model. We found that a block size of $40 \times 40 \times t$, where $t$ is the number of frames in the video, is suitable, as it is large enough to sufficiently occlude the mouth region. After each iteration, the block is displaced by one pixel, and the probability of predicting the correct class is recorded for each occluded pixel. Following this process, a heatmap can be created by averaging the probabilities at each pixel location. The heatmaps are finally normalised and overlaid on the first frame of the video.

\begin{figure*}
\centering
  \centerline{\includegraphics[width=\linewidth]{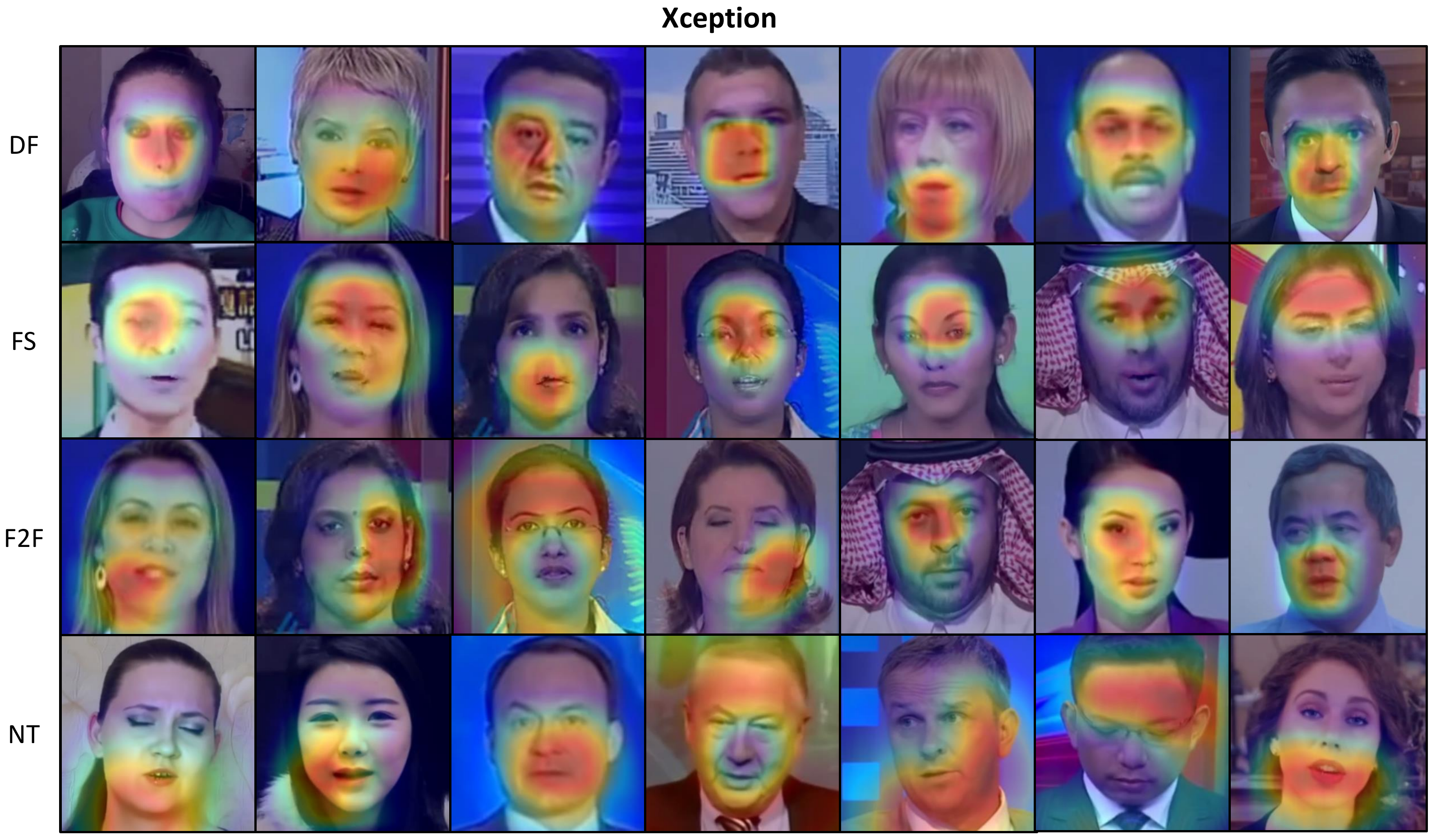}}
\caption{\textbf{Visualisation examples for Xception.} We show examples for Deepfakes (DF), FaceSwap (FS), Face2Face (F2F), and NeuralTextures (NT).}
 \label{fig:viz_xception}
\end{figure*}

\begin{figure*}
\centering
  \centerline{\includegraphics[width=\linewidth]{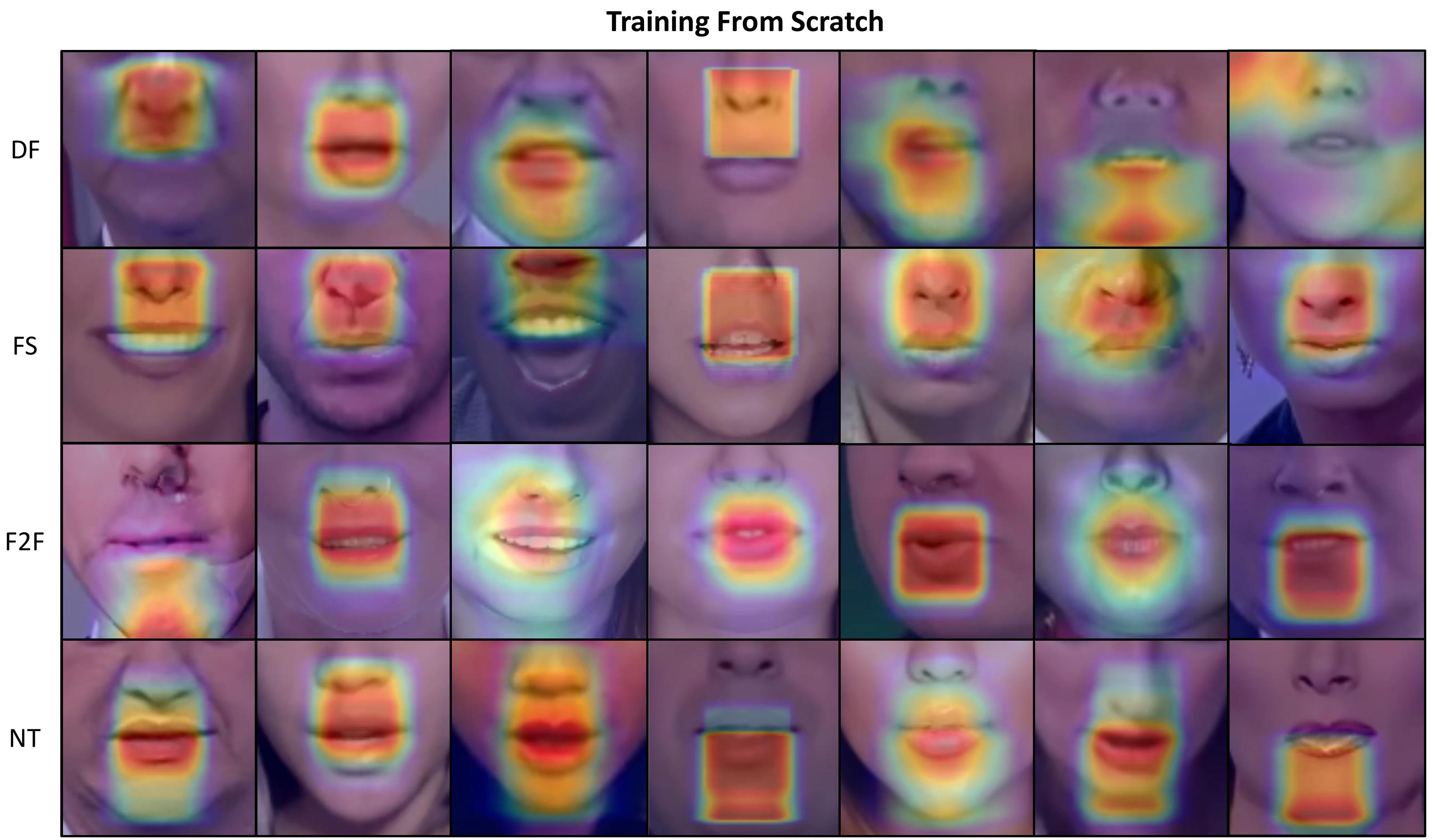}}
\caption{\textbf{Visualisation examples for spatiotemporal network without lipreading pretraining.} We show examples for Deepfakes (DF), FaceSwap (FS), Face2Face (F2F), and NeuralTextures (NT).}
 \label{fig:viz_scratch}
\end{figure*}

\begin{figure*}
\centering
  \centerline{\includegraphics[width=\linewidth]{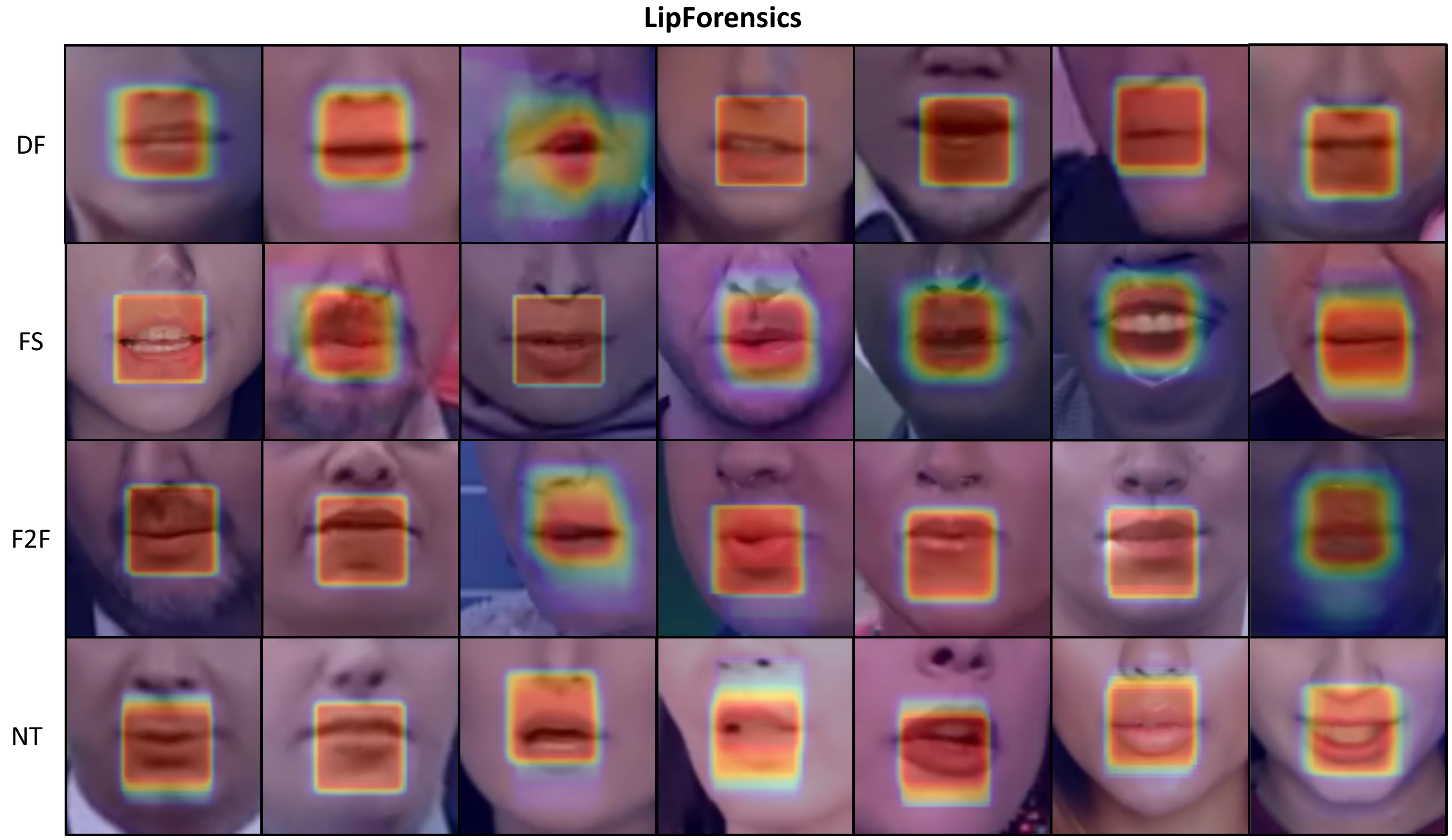}}
\caption{\textbf{Visualisation examples for LipForensics.} We show examples for Deepfakes (DF), FaceSwap (FS), Face2Face (F2F), and NeuralTextures (NT).}
 \label{fig:viz_lipforensics}
\end{figure*}

We show visualisation examples for Xception \cite{rossler2019faceforensics++} (see Figure \ref{fig:viz_xception}) as well as for training the spatiotemporal network from scratch (see Figure \ref{fig:viz_scratch}) and LipForensics (see Figure \ref{fig:viz_lipforensics}). As mentioned in the main text, unlike Xception, LipForensics consistently relies on the mouth region. Interestingly, without lipreading pretraining, the network often seems to rely on regions other than the mouth (such as the nose), despite the (conservative) mouth crop. This is more the case for the face swapping methods, Deepfakes and FaceSwap.

\end{document}